\documentclass[conference]{IEEEtran}
\IEEEoverridecommandlockouts
\usepackage{graphicx}  

\usepackage{multirow}
\usepackage{cite}
\usepackage{multirow,array}
\usepackage[left=0.71in,top=0.94in,right=0.71in,bottom=1.18in]{geometry}
\usepackage{url}
\usepackage{algorithm}
\usepackage{algorithmicx}
\usepackage{algpseudocode}
\usepackage{amsmath}
\usepackage{booktabs}
\newcommand{\degree}{^\circ}
\begin{document}
	\title{\huge Tactical Reward Shaping: Bypassing\\
		Reinforcement Learning with Strategy-Based Goals 
		\footnoterule\thanks{$^{*}$This work was supported by the National Natural Science Foundation of China, project number 61850410527, and the Shanghai Young Oriental Scholars project number 0830000081}}
	
	\author{\authorblockN{ Yizheng Zhang and Andre Rosendo}
		\authorblockA{\textit{Living Machine Lab}\\
			\textit{Shanghaitech University}\\
			\textit{393 Mid HuaxiaRoad
				Pudong District, Shanghai, China}\\
			\textit{\{zhangyzh1, arosendo\}@shanghaitech.edu.cn}\\}}%
	
	\maketitle
	\begin{abstract}
		Deep Reinforcement Learning (DRL) has shown its promising capabilities to learn optimal policies directly from trial and error. However, learning can be hindered if the goal of the learning, defined by the reward function, is ''not optimal". We demonstrate that by setting the goal/target of competition in a counter-intuitive but intelligent way, instead of heuristically trying solutions through many hours the DRL simulation can quickly converge into a winning strategy. The ICRA-DJI RoboMaster AI Challenge is a game of cooperation and competition between robots in a partially observable environment, quite similar to the Counter-Strike game. Unlike the traditional approach to games, where the reward is given at winning the match or hitting the enemy, our DRL algorithm rewards our robots when in a geometric-strategic advantage, which implicitly increases the winning chances. Furthermore, we use Deep Q Learning (DQL) to generate multi-agent paths for moving, which improves the cooperation between two robots by avoiding the collision. Finally, we implement a variant A* algorithm with the same implicit geometric goal as DQL and compare results. We conclude that a well-set goal can put in question the need for learning algorithms, with geometric-based searches outperforming DQL in many orders of magnitude.

	\end{abstract}
	
	
	\section{Introduction}
	The ICRA-DJI Robomaster AI Challenge is an important student robot competition as it requires teams to showcase a well-rounded Computer Science skillets. Self-localization to place all robots in a known map, path planning to guide the movements of our robot, computer vision for enemy detection, autonomous re-supplying of projectiles, and an intelligent decision making are the five main problems which, if solved, will enable a team to win the challenge. There are already exist several frameworks solving the first four problems, thus we are focusing on the decision-making problem to bring intelligence to the robots.  Reinforcement learning has demonstrated its compelling potential in this area. Mnih et al. \cite{mnih2015human} combine reinforcement learning with a deep neural network, the experience replay and fixed Q-targets mechanism, which achieves human-level control on Atari games. Wang et al.\cite{pmlr-v48-wangf16} propose a dueling network architecture to solve the over-estimate problem in DQL. 
	
	However, their work is focused on single-agent reinforcement learning.
	The decision-making problem here contains multi-robot cooperation and competition, which requires agents to learn to interact with others in a shared environment. In this multi-agent reinforcement learning problem (MARL)\cite{bu2008comprehensive}\cite{shoham2007if}\cite{tuyls2012multiagent}, if each agent treats its experience as part of its (non-stationary) environment which means an agent regards other agents as its environment, the policy it learned during training can fail to sufficiently generalize during execution. 
	Lanctot et al.\cite{lanctot2017unified} propose an algorithm using deep reinforcement learning and empirical game-theoretic analysis to compute new meta-strategy distributions.
	On the other hand, Lowe et al.\cite{lowe2017multi} take action policies of other agents into considering and present an adaptation of actor-critic methods that can successfully learn policies that require complex multi-agent coordination. 
	
	The work of \cite{sethy2015real} focuses on the reward function design and makes the agent learn from interaction with any opponents and quickly change the strategy according to the opponents in the BattleCity Game which is quite similar with the ICRA-DJI RoboMaster AI Challenge. But it only includes competition, not cooperation.
	Hong et al. \cite{hong2018deep}present DPIQN and DRPIQN that enable an agent to collaborate or compete with the others in a Multi-Agent System(MAS) by using only high-dimensional raw observations. Srinivasan et al. \cite{srinivasan2018actor} discuss several update rules for actor-critic algorithms in multi-agent reinforcement learning that can work well in zero-sum imperfect information games.
	
	In our work, we use the grid world \cite{1606.01540}\cite{gym_minigrid} and Deep Q Learning baseline\cite{stable-baselines} to build a simulation environment and train policies to control two robots to attack the enemies robots, respectively. The reward is designed through the stag hunt game theory to encourage cooperation. Finally, we found that if the target of the game is set properly, a traditional algorithm such as A* can achieve a better performance than complex reinforcement learning.
	The next section give the problem definition, while the Sections \ref{section:methods} and \ref{section:experiments} introduce our DQL and variant A* algorithm and our experimental results. Finally, the paper makes discusses our results and concludes in Section \ref{section:discuss}.
	
	

	\section{Problem Definition}
	\label{section:ProblemDefinition}
	Even though the AI Challenge environment is partially observable, 
	benefit from excellent Lidar-based Enemy Detection sensing technology that helps us know the enemies' position, we can consider the Partially Observable Markov Decision Process (POMDP) problem as a Markov Decision Process (MDP).
	The MDP is composed of states, actions, transitions, rewards and policy, which were represented by a tuple$<S, A, T, R,\pi>$ respectively.
	\begin{figure}[t!]
		\centering
		\includegraphics[width=0.47\textwidth]{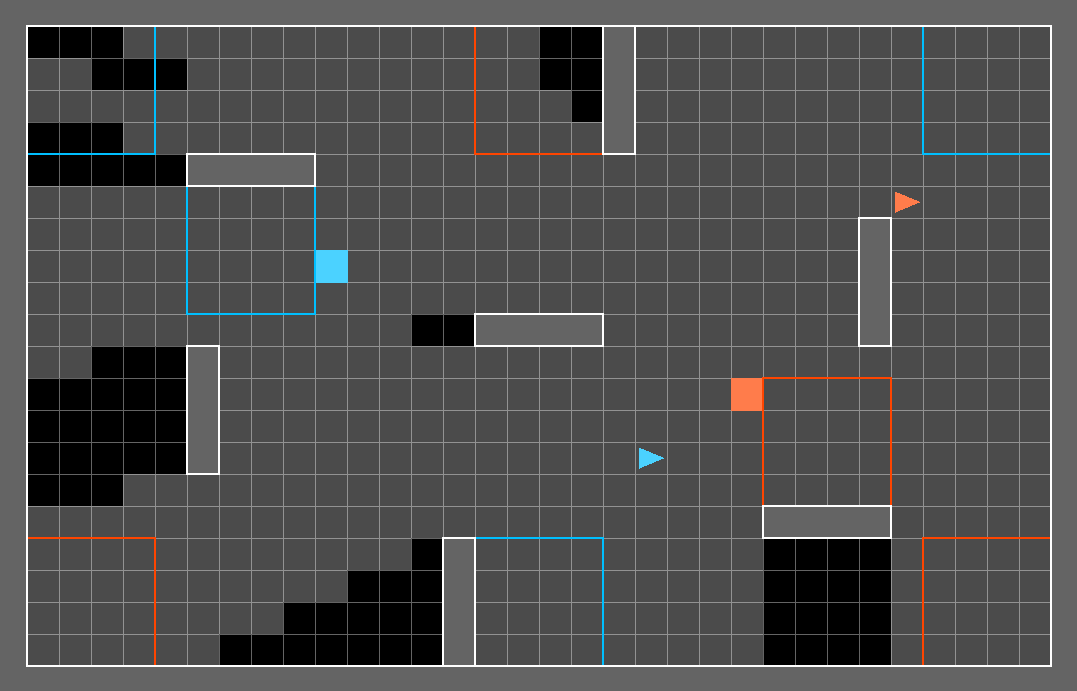}
		\caption{The red and blue colors represent our robots and enemy's robot respectively. The square and triangle are used to distinguish \textnormal{robot$_1$} and \textnormal{robot$_2$}. The grey cuboids with white frames indicate the obstacles in the arena. The black grids are the places that the robot can not see. The fusion view of two robots is represented as gray gird.}
		\label{pomdp2mdp}
	\end{figure}
	
	\begin{itemize}
		\item \textbf{State:} $S$ is the state space which can be discrete or continuous.
		\item \textbf{Action:}  $A$ is the action space which contains a set of discrete actions that an agent can take. The set of actions is state-dependent,
		denoted as $A(s)$.
		\item \textbf{Transition:} $T(s,a)$ is the state transition function $s^{\prime} =T(s,a) $ 
		gives the environment's transition probability to state $s^{\prime}$ on taking action $a$ in $s$. In our simulated environment, it is deterministic.
		\item \textbf{Reward:} $R(s,a)$ is the immediate reward when taking action $a$ at state $s$.
		\item \textbf{Policy:} $\pi(a|s)$ is a probability distribution of all actions under the state $s$, which describes the agent's behaviors. It will be optimized to obtain a higher accumulated reward.
		
	\end{itemize}
	

	Since all the teams buy the same robots from DJI, we can assume the performance of each robot is the same, which means if one robot and another robot are attacking each other, they have the same health points, cause the same damage, and die at the same time. So our robots need to learn to cooperate to attack the enemies. 
	
	\section{Methods}
	\label{section:methods}
	
	\subsection{Lidar-based Enemy Detection}
	\label{subsection:lidar}
	
	\begin{figure}[t!]
		\centering
		\includegraphics[width=0.47\textwidth]{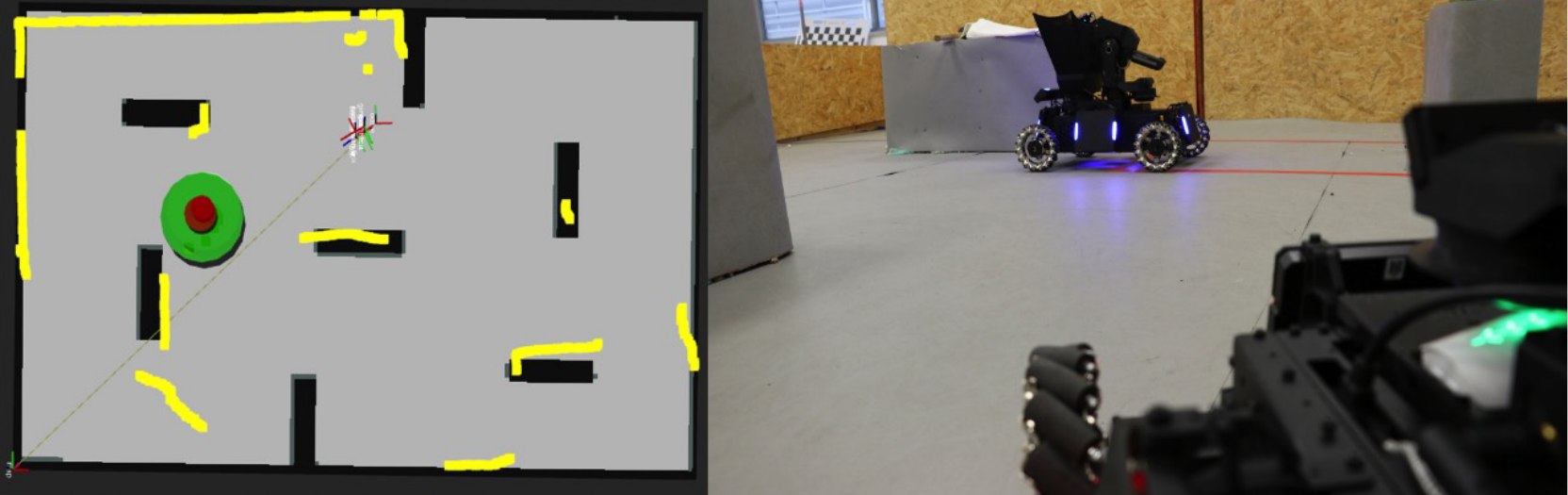}
		\caption{Enemy detected by LiDAR. Our algorithm filters disturbances from our prior knowledge of the map to detect enemies. The left half of the picture shows our detection result. The right half of the picture shows the current situation of two robots in the arena. The green-red concentric circle represents the locations of the enemy robot detected using lidar. The blue circle represents our robot. The yellow lines represent the lidar scan data.
		}
		\label{lidar-detection}
	\end{figure}
	
	We use 2d obstacle detection algorithm to extract obstacles information from lidar data \cite{obstacle_detector}. 
	This algorithm firstly extracts a line from raw lidar data, regarding this line as the secant line inside a circle. We treat this circle as an obstacle that corresponding to an enemy robot or an ally robot.
	However, when a line is generated from the lidar data representing the wall in the environment, it can make a mistake by regarding the wall as a robot.
	To solve this problem, we add a filter to the results we got. Since we know the map, we know where the walls are. If the center of a circle is inside a wall, we filter out this circle. 
	Since our robots can communicate with each other, it is easy to fuse the view of two robots, which helps to distinguish which circle is the ally robot, which circle is the enemy robot.

	\subsection{Stag Hunt Strategy}
	Stag hunt is a game that
	each player can individually choose to hunt a stag or hunt a hare.  If an individual hunts a stag, they must have the cooperation of their partner to succeed. An individual can get a hare by himself, but a hare is worth less than a stag. In our case, we can regard our two robots are the players and the enemies are the stag and the hare. Then we can formulate our payoff table as follow. If two agents are attacking the same enemy which is the stag, two agents obtain the max reward.  The reward function of reinforcement learning can be designed according to this payoff table to encourage cooperation.
	\begin{table}[h]
		\renewcommand{\arraystretch}{1.2}
		\renewcommand{\thefootnote}{\alph{footnote}}
		\caption{Payoff table of stag hunt} \label{staghunttable}
		\begin{minipage} {0.5\textwidth}
			\begin{center}
				\begin{tabular}{*{4}{c|}}
					\multicolumn{2}{c}{} & \multicolumn{2}{c}{\textit{Agent$_1$}}\\\cline{3-4}
					\multicolumn{1}{c}{} &  &  \textit{Enemy$_1$(stag)}  &  \textit{Enemy$_2$(hare)} \\\cline{2-4}
					\multirow{2}*{\textit{Agent$_2$}}  & \textit{Enemy$_1$(stag}  & $(3,3)$ & $(0,2)$ \\\cline{2-4}
					& \textit{Enemy$_2$(hare)}  & $(2,0)$ & $(1,1)$ \\\cline{2-4}
				\end{tabular}
			\end{center}
		\end{minipage}
	\end{table}
	
	\subsection{ Deep Q Learning With Stag Hunt}
	\begin{figure}[t!]
		\centering
		\includegraphics[width=0.47\textwidth]{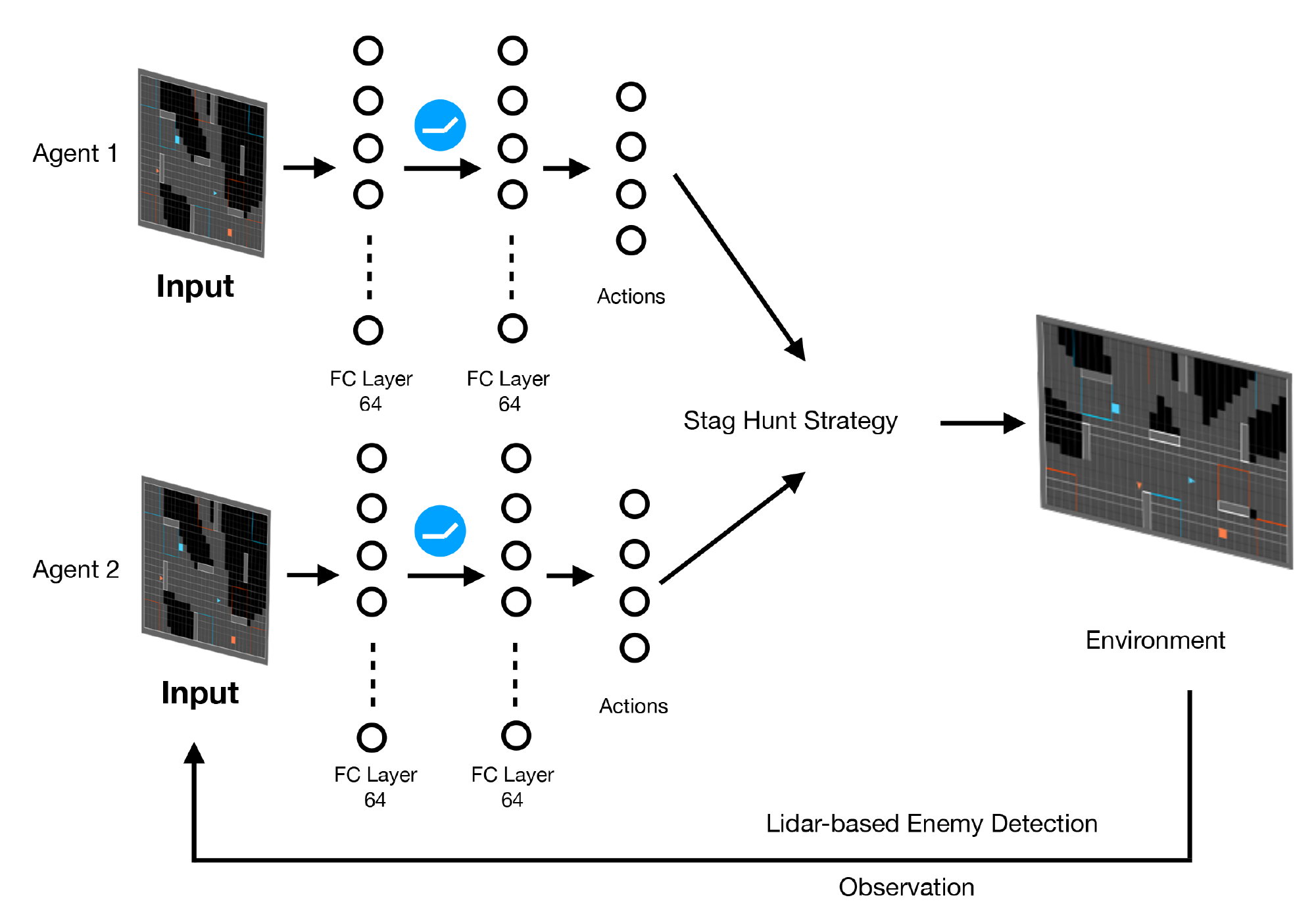}
		\caption{The neural structure of multi-agent Deep Q Learning with reward according to the stag hunt strategy.  Every agent has their network but shares the same structure and parameters. Decision-making processes happen independently and simultaneously. }
		\label{network1}
	\end{figure}
	\subsubsection{Environment}  
	The custom simulation environment is based on a $32\times20$ cell grid world \cite{gym_minigrid} to approximate the real competition venue. Each grid cell is assigned a static class (wall, empty, robot). The agent can read the type of any grid cell within its sensor FOV(RPLIDAR A3: $360 \degree$ horizontal, $32-$cell radial range), which means that it can see any cell that is not obscured by obstacles. Taking the advantage of the excellent characteristics of the sensor and combine the lidar-based enemy detection of two robots, we can approximately know the enemies' coordinates at any time actually, which means we can consider the problem as an MDP instead of POMDP.
	
	Since the position of the obstacles(the wall in this competition) will not change, to simplify the observation, we set observation as the positions of the four robots(agent, ally, enemy1,enemy2). And use the last seen position as coordinates if the enemy is not in sight. 
	
	To distinguish these 4 robots concisely, we have the following notation:
	\begin{itemize}
		\item \textit{agent$_1$:} one of our robot.
		\item \textit{agent$_2$:} the other one of our robot.
		\item \textit{enemy$_1$:} one of the enemy robot.
		\item \textit{enemy$_2$:} the other one of the enemy robot.
		\item \textit{stag:} the enemy robot we want to attack, selected from \textit{enemy$_1$} and \textit{enemy$_2$}
		\item \textnormal{hare:} the enemy robot we want to avoid and not attack, selected from \textit{enemy$_1$} and \textit{enemy$_2$}
	\end{itemize}

	\subsubsection{Action Space} The action space consists of five discrete actions: \textit{Up, Down, Left, Right and Stop}. 
	\subsubsection{State Space}
	Since the environment is stationary, the only moving objects are four robots, so we set the coordinate of four robots as the state space.
	\subsubsection{Network Structure}
	Two agents have shared the same network structure and parameters. 
	The network structure is illustrated in Fig. \ref{network1}. The network receives the state vector which is an 8 dimensions vector as the input. Since our input is not a figure, there is no need to use the convolution layer to extract features. The first three hidden layers are fully-connected layers with the relu function as the activation function. The final outputs layer to the action distribution is connected with another fully-connected layer with the action space size.

	\subsubsection{Reward Function}
	Each robot has a referee system to show their health, and when we attack, the blood volume will decrease, ideally, this can be treated as a good reward to direct the robot attack enemy. However, monitoring their blood volume in real-time is a very big challenge for the computer vision algorithm. 
	
	According to the actual situation, the farther the distance, the worse the accuracy of the shooting. To simplify the shooting function of the robot, we set up a range, and when the enemy is within that range, we think it's under attack.  So that if the enemy is within the range of our robot, which means our robots are attacking the enemy robot, our robots receive a high reward. 
	
	
	However, giving rewards according to whether the enemy robot is within our robot range leads to very low efficient learning because the number of successes is too small that it is difficult to learn useful things from these successes. 
	
	So we modified the reward that it is given corresponding to the distance between \textit{agent$_1$} and \textit{stag}.
	The reward is given as follow:
	$${{ r_1} = -\frac{{\rm distance}_{ \mathit{agent_1}-\mathit{stag}  } }
		{arena.length+ arena.width}}$$


	One step closer, 
	while our robot is moving toward the \textit{stag}, we don't want to be attacked by another robot, so it requires us to plan the path to avoid another robot.
	Based on $ r_1$, we add another item:
	$$ punishment = -\frac{\rm{distance}_{\mathit{agent_1}-\mathit{hare}}}{attackRange}$$
	So the new reward is given as follow:
	$$ r_2 = r_1+punishment$$
	
	
	With this reward $ r_2$, the agent can achieve the goal that goes to attack \textit{enemy$_1$} and avoid \textit{enemy$_2$}'s attack.
	
	\textit{Agent$_1$} and \textit{agent$_2$} share the same parameters. And the reward will increase if $2$ robots go to attack the same enemy at the same time to encourage cooperation according to the stag hunt strategy.
	
	\subsection{Variant A* algorithm for path planning}
	The A* algorithm has been existed for half a century and widely used in path finding and graph traversal. The original A* algorithm can find the shortest path from $agent_1$ to $stag$ in the grid map. To meet our requirements, we made some modifications. First, we discard the last few points to the $stag$ because we only want the $stag$ is within the attack range of $agent_1$ but not reach the $stag$'s position. Second, we set a safe distance to the $hare$ to avoid been attacked while moving towards $stag$. The second point is quite the same as the punishment item in $ r_2$.


	\section{Experimental Results}
	\label{section:experiments}
	\subsection{Experimental Setup}
	\begin{figure}[t!]
		\centering
		\includegraphics[width=0.47\textwidth]{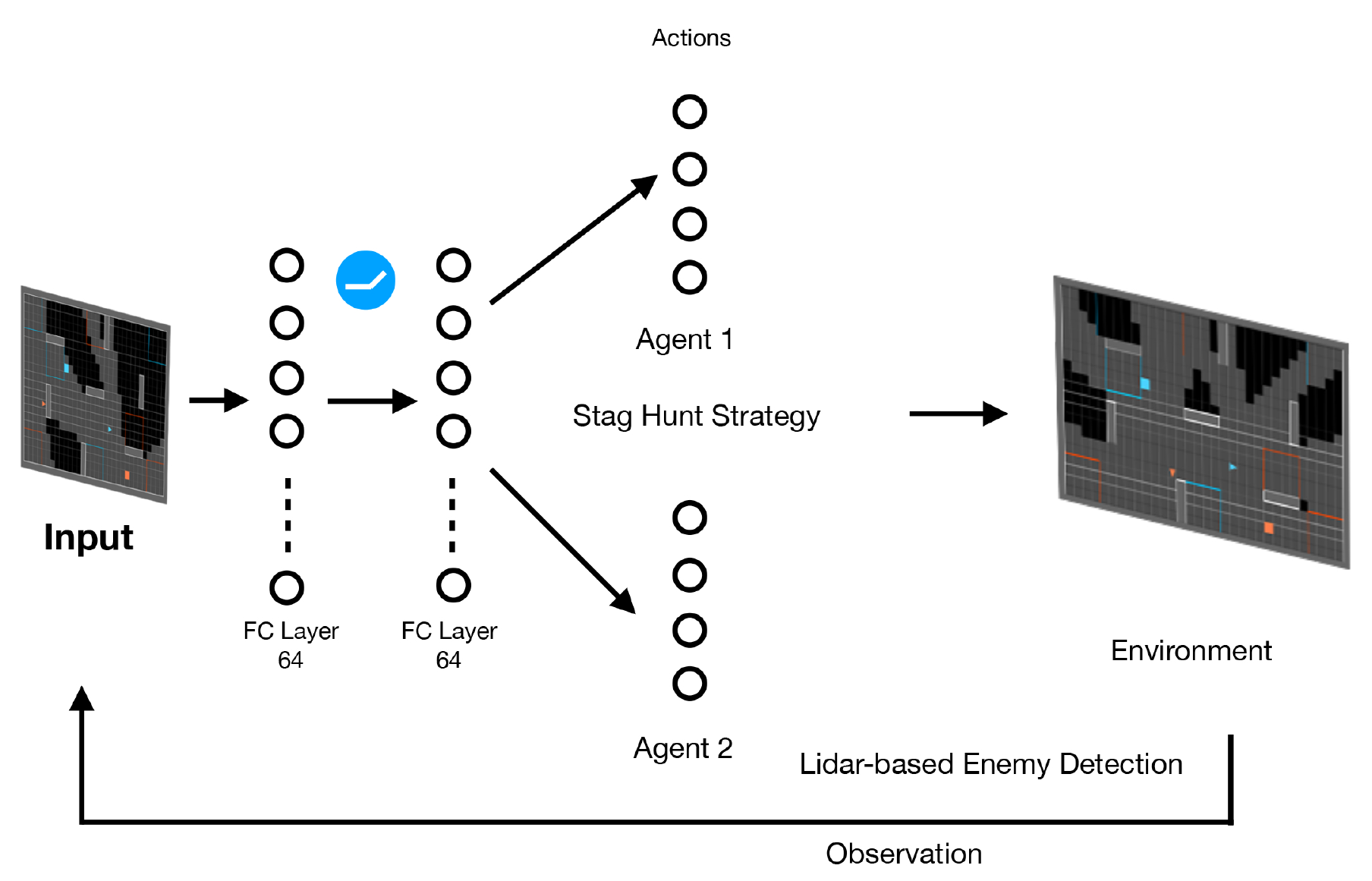}
		\caption{The neural structure of multi-agent Deep Q Learning with reward according to the stag hunt strategy. One network controls two agents' behavior at the same time.}
		\label{network2}
	\end{figure}
	\subsubsection{Hyper parameters}
	the discount factor $\gamma$ is set to $0.99$ to enable the agent a long term view. The frequency of the target network updates every $1000$ episode. We also use a prioritized replay technique to speed up training speed and the prioritized replay $\alpha$ is $0.6$. The exploration fraction parameter is set to $0.8$ initially and it will be linearly decreased gradually during the learning process until $0.3$. The learning rate of the network is set to $0.01$. We total training episode number is $2,000,000$ and the replay buffer's size is $1,000,000$.
	\subsubsection{Network Models}
	we test three kinds of models to see which perform better. 
	\begin{itemize}
		\item Model $1$: $2$ DQNs share the same parameters. The dense reward is given at each step according to the distance from the target enemy. At the training period, the reward obtained will increase if the robots meet the stag-stag case to encourage cooperation. The structure is shown in Fig. \ref{network1}.
		
		\item Model $2$: Since \textit{agent$_1$} and \textit{agent$_2$}  have the same goal and there is no conflict between them, that means we can use $1$ DQN to controls two agents at the same time.  One agent has 5 actions, so this network's action space is grown to 25. The structure is shown in Fig. \ref{network2}. It is also given a dense reward. 
		
		\item Model $3$: $2$ DQNs share the same parameters. The sparse reward is given only when the robot can attack the enemy and reward will also increase when meeting the stag-stag case.
	\end{itemize}
	
	\subsubsection{Enemy robot}
	the enemy robot is generated randomly on the map in every episode. 
	
	
	\begin{figure}[t!]
		\centering
		\includegraphics[width=0.47\textwidth]{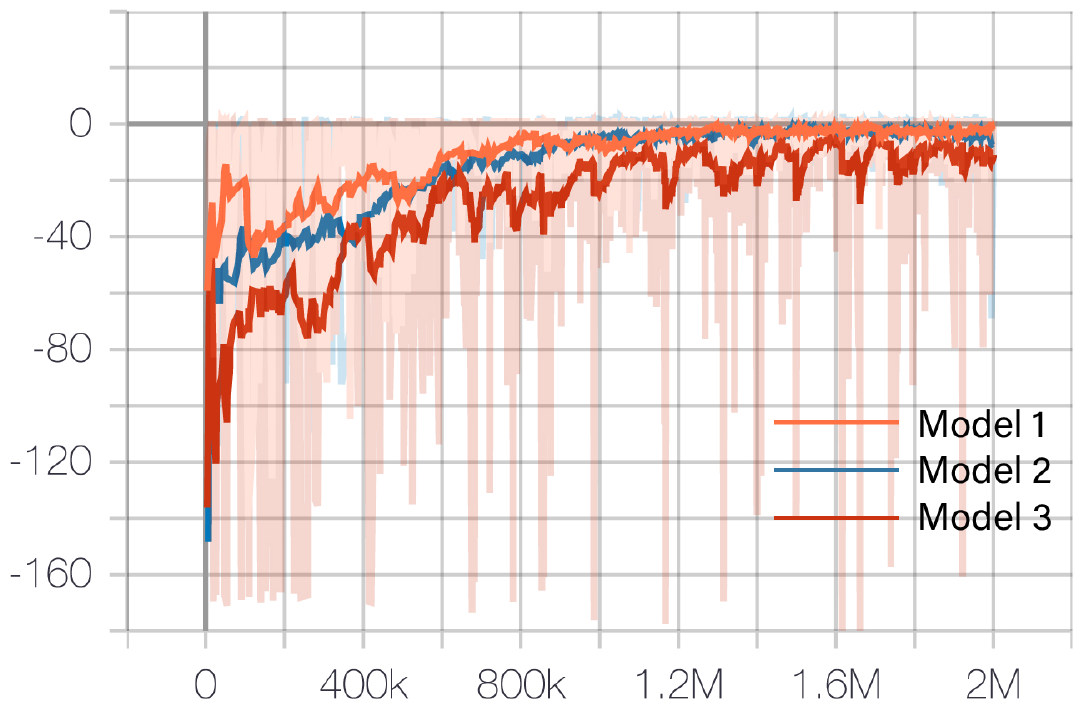}
		\caption{The mean episode reward of the training.}
		\label{episode_reward}
	\end{figure}
	
	\begin{figure}[t!]
		\centering
		\includegraphics[width=0.47\textwidth]{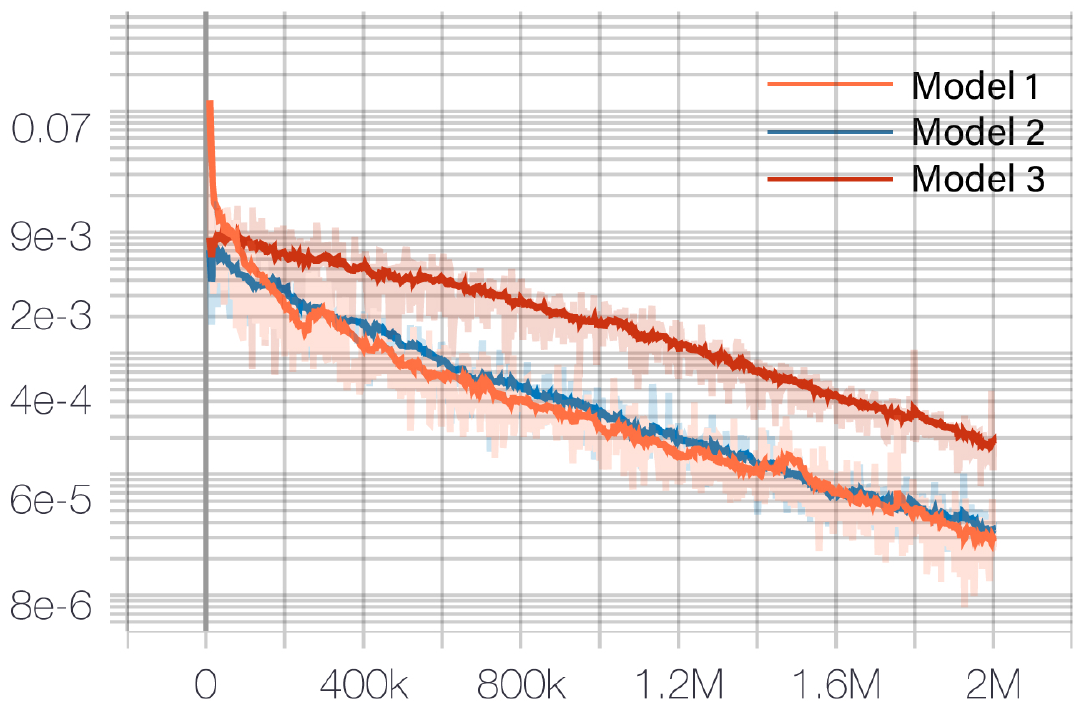}
		\caption{The mean loss of the training.}
		\label{loss_loss}
	\end{figure}
	
	\begin{figure}[t!]
		\centering
		\includegraphics[width=0.47\textwidth]{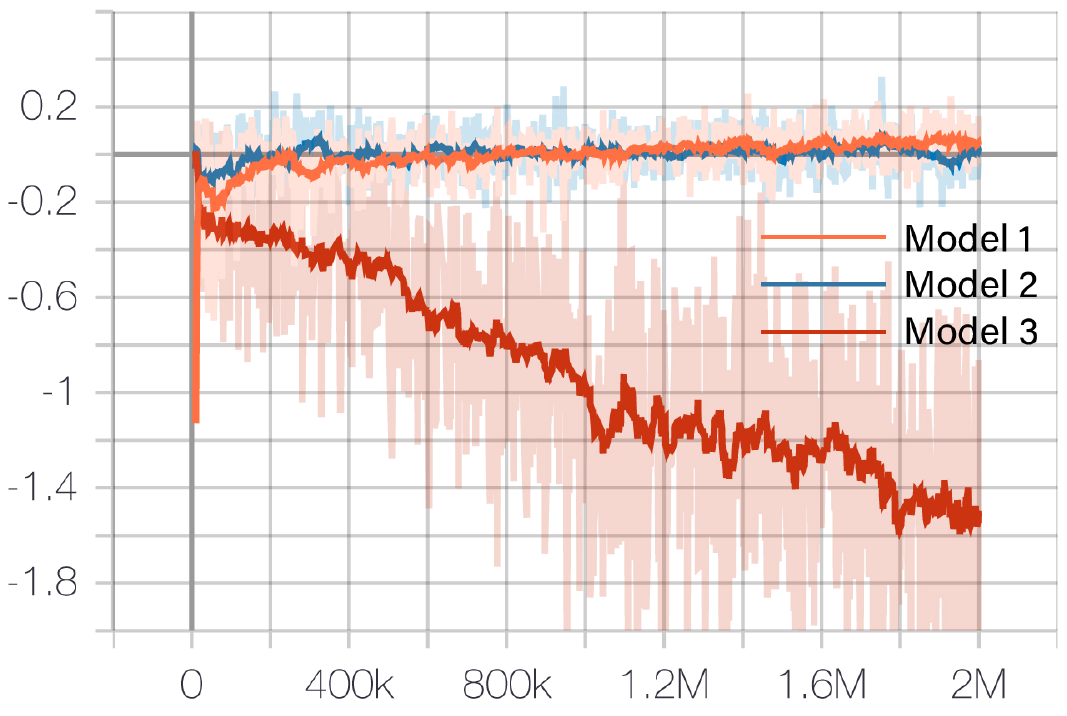}
		\caption{The mean temporal difference error of the training.}
		\label{loss_td_error}
	\end{figure}
	
	\subsection{Training}
	
	The simulation results of training are shown on Fig. \ref{episode_reward}, Fig. \ref{loss_loss} and Fig. \ref{loss_td_error}.
	We can see model 1 can reach the highest reward in three models from Fig. \ref{episode_reward}. Model 2 achieves similar performance to Model 1 after $1.2$ million episodes and Mode 3's performance is a little lower than the other two which means sparse reward makes learning more difficult. 
	Model 1 and Model 2 have a similar loss and temporal difference error curves which means these two models can achieve similar results.
	Model 3's loss decreases slower than Model 1's and the temporal difference error also increases which makes sense due to the sparse reward. According to the training results, we choose Model 1 as our final Model.

	\begin{figure}[t!]
		\centering
		\includegraphics[width=0.47\textwidth]{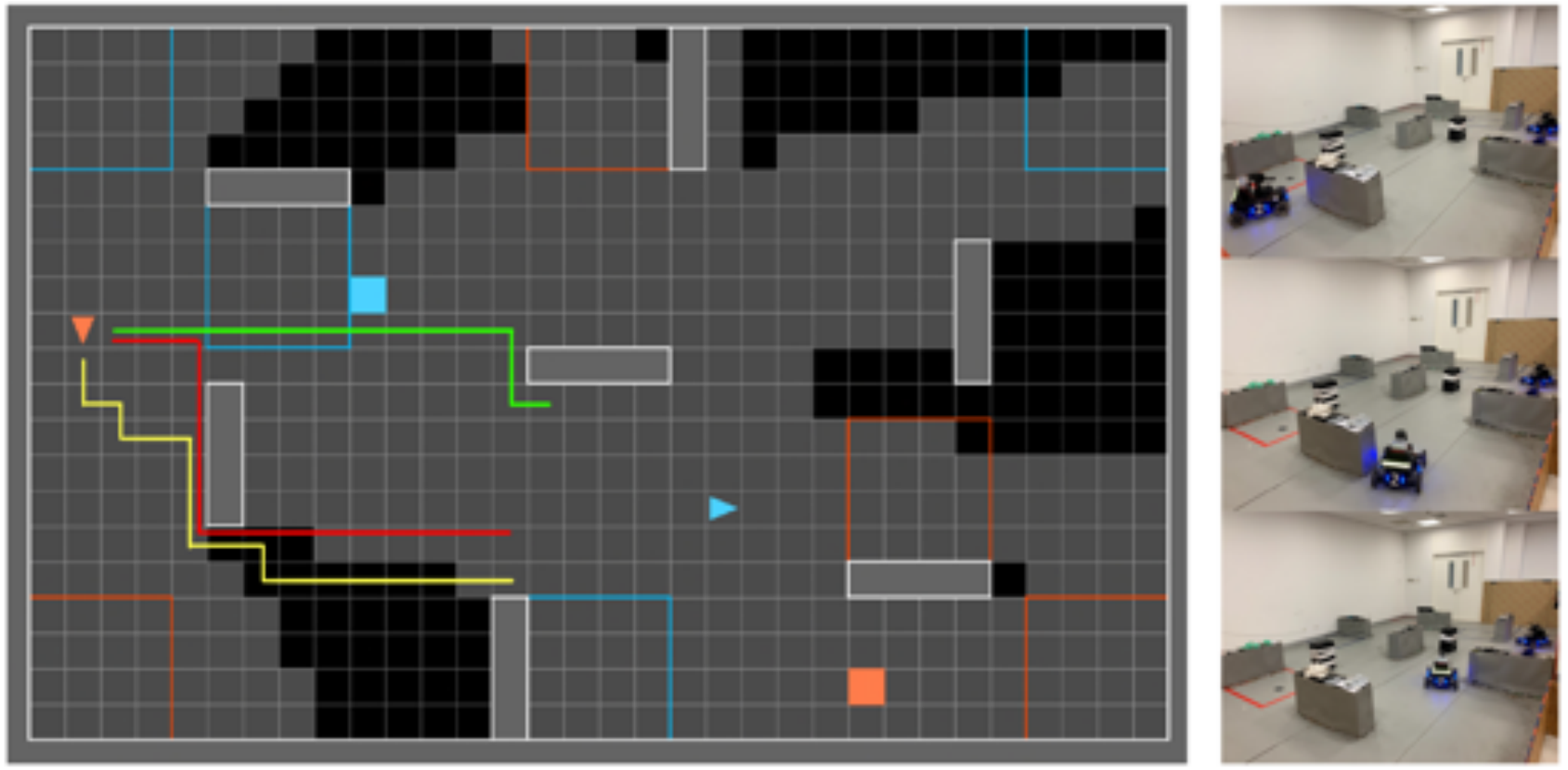}
		\caption{The green path and yellow path are corresponding to reward function $r_1$ and $r_2$. The red path is generated by the variant A* algorithm. In the set of pictures on the right, two black boxes represent enemy robots. We implemented the learned policy from the simulator to the real robots in our arena. The learned policy serves as a global planner and gives the next target position. }
		\label{pathcompare1}
	\end{figure}
	\subsection{The performance with different reward function}
	\subsubsection{Reward function $ r_1$}
	To direct the robot to move toward the enemy robot, the reward function need to have the property that the shorter the distance between two robots is, the higher the reward it is. So we design the  reward function as follow:
	$$r_1 = -\frac{distance_{\mathit{agent}-\mathit{stag}}}{\beta}$$
	
	Where $\beta$ is the normalization factor and is set to
	$arena.length+ arena.width$.
	
	After training with $r_1$, the DQL will find the shortest path to achieve the target enemy robot. The green path in Fig. \ref{pathcompare1} shows the path generated under such a reward function.
	
	\subsubsection{Reward function $r_2$}
	As we can see the green path from Fig. \ref{pathcompare1}, it will be very close to the rectangle enemy robot and being attacked while it's moving. So a punishment item is added to $r_1$ to encourage avoiding another enemy robot while moving. The punishment item is given as:
	$$punishment = -\frac{\rm{distance}_{\mathit{agent}-\mathit{hare}}}{attackRange}$$

	The yellow path in Fig. \ref{pathcompare1} shows the path generated under such a reward function. It is a safer path than the green path and more practical in the competition.


	\subsection{Planning using variant A* algorithm}
	Since the original A* algorithm will find the shortest path in the grid map and the DQL will also find the optimal path after millions of trials, the performance of these two algorithms actually will become very similar. The green path can also be obtained by the original A* algorithm in this situation, which does not meet our requirement. After adding the safe distance function to the original A* algorithm, the variant A* algorithm can find a path that can also avoid the other enemy robot. It is shown as the red path in the  \ref{pathcompare1}.
	\subsection{Comparison between variant A* algorithm and DQL}
\begin{figure}[t!]
		\centering
		\includegraphics[width=0.47\textwidth]{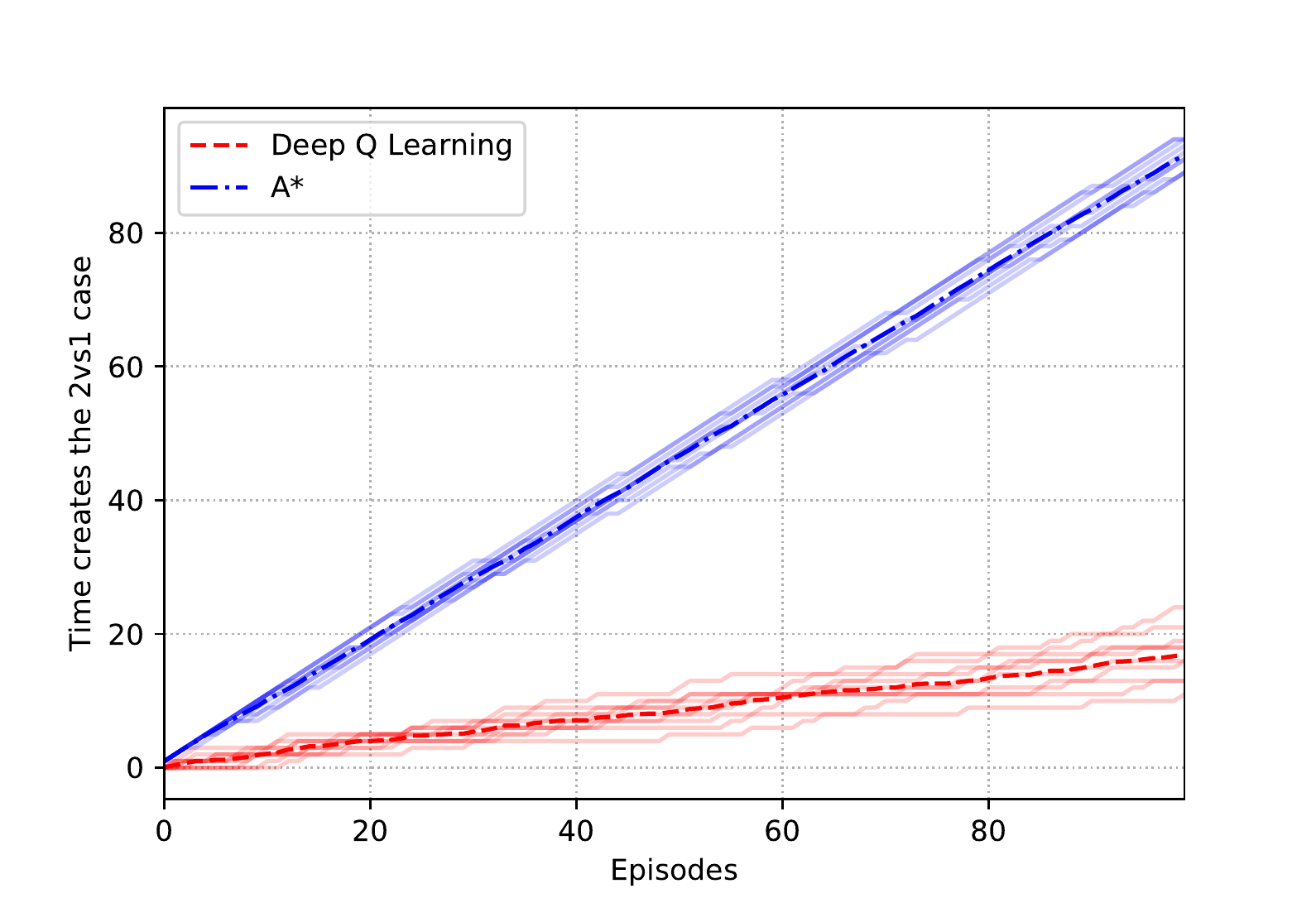}
		\caption{The green path and yellow path are corresponding to reward function $r_1$ and $r_2$. The red path is generated by variant A* algorithm.}
		\label{compare}
	\end{figure}

	To evaluate these two algorithms, we count how many times can they create the 2 v.s. 1 case that can increase our chance to win. So the blue robots are implemented with the variant A* algorithm and the red robots use the trained Deep Q Network from Model 1. Four robots are generated randomly in the arena. Once one team creates the 2 v.s. 1 case, this match is over. We counted the number of successes per team after 100 matchups. And repeat the whole process several times to eliminate the potential impact of randomness.
	The result is shown in Fig. \ref{compare}.

\begin{table}[t]
	\renewcommand{\arraystretch}{1.2}
	\renewcommand{\thefootnote}{\alph{footnote}}
	\caption{Successful rate of creating the 2vs 1 case} \label{ratetable}
	\begin{minipage} {0.5\textwidth}
		\begin{center}
\begin{tabular}{cc}
	\toprule  
	 Algorithm&  Successful rate \\
	\midrule  
	Deep Q Learning& $13\%-$24\%\\
	A*&$89\%-$94\% \\
	\bottomrule 
\end{tabular}
		\end{center}
	\end{minipage}
\end{table}

	\section{Discussion and Conclusions}
	\label{section:discuss}
	The ICRA-DJI RoboMaster AI Challenge includes a variety of major robotics technologies. Each function, such as self-localization, has its noise because the sensor is not noise-free. 
	
	The environment turns out to be a POMDP.  A lot of teams only use the camera to detect the enemy robot which limits the robot's perception due to the small FOV of the camera. We firstly develop the lidar-based enemy detection technique that enhances the robot's perception capability and turns the POMDP problem into an MDP problem.
	
	And each team will manual design their strategy according to their understanding of the rules and try to take advantage of these rules. Due to different team strategies, it is also difficult to ensure that the strategy is effective for the opponent and win the game.
	
	As presented in \cite{1241933}\cite{6707164}, they try to win the game by
	estimating the enemy's strategy and then adjust their strategy according to the estimated result. It can be difficult to generalize this solution to different opponents. However, we found out that a well-defined target can simplify the problem that can be solved even without the need of reinforcement learning.
	
	Unlike this approach, we are not focusing on a general strategy that can win the game. Our approach focuses on the strategy which is derived from the stag hunt game that can increase our chances of being in a position of advantage. We gave a different payoff in different situations and obtained the stag hunt payoff table \ref{staghunttable}. According to this, we find that creating 2 vs 1 scenario can increase our reward and winning chance. 
	
	Following the goal of reaching a 2 vs 1 scenario that implicitly tries to create a geometric-strategic advantage, we use DQL and the variant A* algorithm to do path planning. For Deep Q Learning, it is a model-free learning algorithm. In our experiment, the enemy robot is treated as a part of the environment and the agent needs millions of trials  ($2,000,000$ episodes) to obtain a good performance.  We test two kinds of networks and evaluate their performance from mean episode reward and loss. Then we choose a structure with two DQNs controlling two robots as our final structure. 
	On the other hand, the variant A* algorithm is derived from the same goal but in a more traditional method. 
	From Fig. \ref{pathcompare1} and Table \ref{ratetable}, 
	it shows that it creates 2 vs 1 scenarios about four times as many times as DQL. Also, the learning method took hours to train while the A* algorithm only need about $100$ milliseconds. Depending on this, we conclude that a well-set, implicit goal can simplify a problem and allow us to use a relatively low-level algorithm to solve a problem that could have required hours of computational time with a learning algorithm.
	
	There is still a huge space for our agents to improve in the future. 
	Our algorithm made decisions according to the position of the opponents at each time step but didn't take the history information into account. We believe that this information can help our decision-making module more intelligent.

	\bibliographystyle{IEEEtran}
	\bibliography{ref}


	
\end{document}